\documentclass{article}

\usepackage{arxiv}

\usepackage[utf8]{inputenc} 
\usepackage[T1]{fontenc}    
\usepackage{hyperref}       
\usepackage{url}            
\usepackage{booktabs}       
\usepackage{amsfonts}       
\usepackage{nicefrac}       
\usepackage{microtype}      
\usepackage{lipsum}		
\usepackage{graphicx}
\usepackage{doi}
\usepackage{subcaption}
\usepackage{multirow}
\usepackage{bigstrut}
\usepackage{xcolor}
\usepackage{cite}
\usepackage{amsmath}
\usepackage{tabularx}
\usepackage{dblfloatfix}
\usepackage{array}

\title{BackLink: Supervised Local Training with Backward Links}

\author{ \href{https://orcid.org/0000-0003-2410-7315}{\includegraphics[scale=0.06]{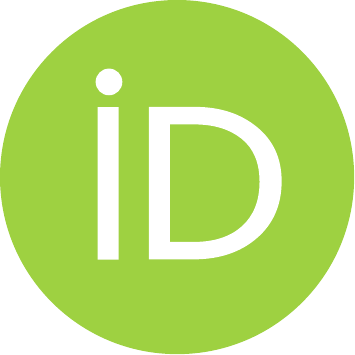}\hspace{1mm}Wenzhe~Guo}\\
	Department of Electrical and Computer Engineering\\
	King Abdullah University of Science and Technology\\
	Thuwal 23955, Saudi Arabia\\
	\texttt{wenzhe.guo@kaust.edu.sa} \\
	\And
	\href{https://orcid.org/0000-0000-0000-0000}{\includegraphics[scale=0.06]{orcid.pdf}\hspace{1mm}Mohammed~E.~Fouda} \\
	Center for Embedded \& Cyber-physical Systems\\ University of California, Irvine,\\
	CA 92612 USA \texttt{ foudam@uci.edu} \\
	\And
	\href{https://orcid.org/0000-0003-1849-083X}{\includegraphics[scale=0.06]{orcid.pdf}\hspace{1mm}Ahmed~M.~Eltawil} \\
	Department of Electrical and Computer Engineering\\
	King Abdullah University of Science and Technology\\
	Thuwal 23955, Saudi Arabia\\
	\texttt{ahmed.eltawil@kaust.edu.sa} \\
	\And
	\href{https://orcid.org/0000-0001-7742-1282}{\includegraphics[scale=0.06]{orcid.pdf}\hspace{1mm}Khaled~N.~Salama} \\
	Department of Electrical and Computer Engineering\\
	King Abdullah University of Science and Technology\\
	Thuwal 23955, Saudi Arabia\\
	\texttt{khaled.salama@kaust.edu.sa} \\
}

\date{}

\begin{document}
\maketitle

\begin{abstract}
Empowered by the backpropagation (BP) algorithm, deep neural networks have dominated the race in solving various cognitive tasks. The restricted training pattern in the standard BP requires end-to-end error propagation, causing large memory cost and prohibiting model parallelization. Existing local training methods aim to resolve the training obstacle by completely cutting off the backward path between modules and isolating their gradients to reduce memory cost and accelerate the training process. These methods prevent errors from flowing between modules and hence information exchange, resulting in inferior performance. This work proposes a novel local training algorithm, BackLink, which introduces inter-module backward dependency and allows errors to flow between modules. The algorithm facilitates information to flow backward along with the network. To preserve the computational advantage of local training, BackLink restricts the error propagation length within the module. Extensive experiments performed in various deep convolutional neural networks demonstrate that our method consistently improves the classification performance of local training algorithms over other methods. For example, in ResNet32 with 16 local modules, our method surpasses the conventional greedy local training method by 4.00\% and a recent work by 1.83\% in accuracy on CIFAR10, respectively. Analysis of computational costs reveals that small overheads are incurred in GPU memory costs and runtime on multiple GPUs. Our method can lead up to a 79\% reduction in memory cost and 52\% in simulation runtime in ResNet110 compared to the standard BP. Therefore, our method could create new opportunities for improving training algorithms towards better efficiency and biological plausibility.
\end{abstract}


\section{Introduction}

{D}{eep} neural networks (DNNs) have achieved great success in solving complex tasks, such as image processing \cite{RN69, RN27}, language processing \cite{RN44, NIPS2017_3f5ee243}, object detection \cite{RN70, 7780460}, and medical diagnostics \cite{RN68, Rn81}. Empowered by the backpropagation (BP) algorithm, DNNs have become the mainstream approach in almost all cognitive applications. However, the standard BP suffers from the well-known backward locking problem that only permits the updates of a module after all dependent modules finish execution of both forward and backward passes \cite{10.5555/3305381.3305549}. This problem arises as errors are propagated backward from the top in a layer-by-layer fashion to update downstream module parameters. It restricts the network to perform training in a sequential manner. Intermediate tensors and operations necessary for module updates must be saved during the forward pass, causing high memory cost and frequent memory access \cite{RN18}. The memory constraints generally impose limitations in the training of the state-of-the-art DNNs on high-resolution inputs and large batch sizes. The strong inter-layer backward dependency prohibits training parallelization, slowing down the training process. This inefficient training process also holds back DNNs from being deployed in resource-constrained platforms such as edge devices.

The major obstacle imposed by backward locking is the heavy dependency between layers due to the error feedback signals. Various local training methods are introduced to tackle this difficulty by cutting off the feedback path \cite{10.5555/3305381.3305549, RN18, RN19, pmlr-v80-huang18b, RN41, pmlr-v119-belilovsky20a, wang2021revisiting, RN76, RN82}. These methods split the network into multiple modules, attach an auxiliary network to each module, and train each module separately and simultaneously or asynchronously. Since training happens locally, intermediate states can be saved in buffers temporally before parameter updates, eliminating the need for memory storage and access. Moreover, local training allows for the parallel execution of the forward and backward pass, significantly accelerating the training process. 

Some works were reported to train networks in a layer-wise cascading fashion \cite{RN19, pmlr-v80-huang18b}. In other words, one layer is fully trained with an auxiliary classifier before moving on to train the next layer. For example, Marquez et al. proposed to train each layer using a multiple-layer fully-connected classifier in a cascading fashion to improve training efficiency \cite{RN19}. However, the performance of these approaches is usually limited to small datasets. Other works focus on designing locally supervised auxiliary networks to produce error feedback signals to train individual network modules synchronously \cite{10.5555/3305381.3305549,  pmlr-v80-huang18b, RN41, pmlr-v119-belilovsky20a, wang2021revisiting, RN76, RN82}. Jaderberg et al. proposed a decoupled neural interface method that synthesizes gradients locally to approximate the true gradients and eliminates the need to receive feedback signals \cite{10.5555/3305381.3305549}. Belilovsky et al. reported using a convolutional classifier to scale the network performance to ImageNet \cite{RN76}. Nokland et al. applied two cla  ssifiers equipped with different loss functions to train each layer \cite{RN41}. Wang et al. analyzed the defect of greedy local training methods from an information perspective and proposed a two-branch auxiliary network aiming to compute information propagation loss \cite{wang2021revisiting}. Significant improvement was demonstrated in various tasks. However, all these works are built on top of the same conventional greedy local training configuration that completely cuts off the backward path between local modules and prevents errors from flowing between them and hence the exchange of information.

This work aims to restore the inter-layer backward dependency and facilitate the information flow to improve network performance while preserving the computational advantages of local training methods. We introduce BackLink, a new local training algorithm that permits the errors generated in the current network module to flow back to its predecessor along a restricted propagation path. We perform extensive experiments to evaluate its effectiveness and analyze the computational costs in terms of memory consumption and computational time.
The main contributions of this work are summarized as follows.
\begin{itemize}
    \item Proposes a novel local training algorithm that introduces inter-module backward dependency to facilitate information flow and improve network performance with small computational overheads.
    \item Extensive experiments demonstrate that the proposed method outperforms the state-of-the-art methods \cite{pmlr-v119-belilovsky20a, wang2021revisiting} in classification performance. Increases in accuracy of up to 3.82\% and 4.05\% are observed on CIFAR10 for ResNet32 and ResNet110, respectively.
    \item Memory analysis and multi-GPU implementation reveal that the proposed method achieves largely reduced memory costs and significant training acceleration over BP. A 79\% reduction in memory cost and a 52\% decrease in runtime can be achieved in ResNet110.
\end{itemize}

\begin{figure*}[!t]
    \centering
    \includegraphics[width=\linewidth]{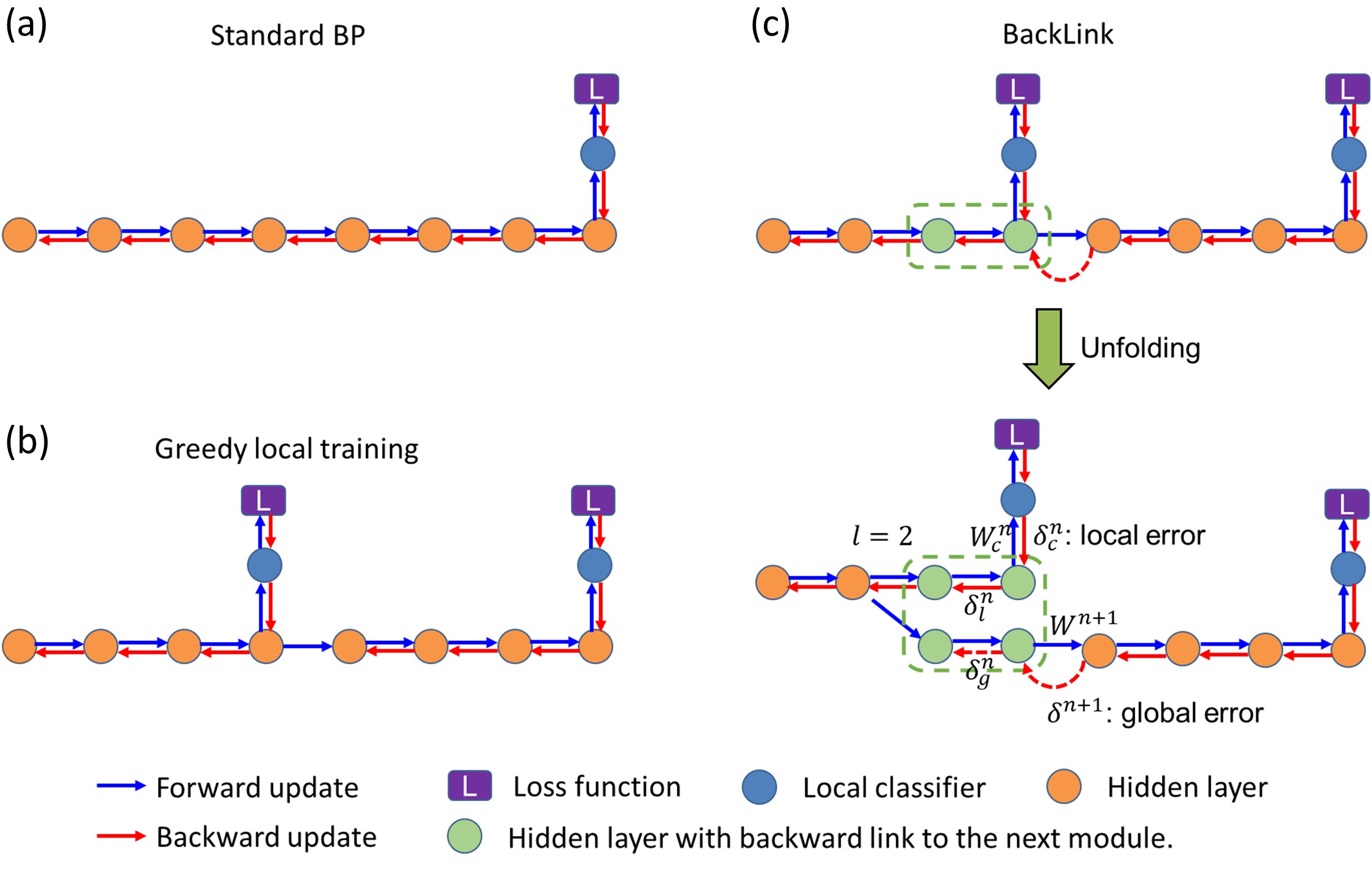}
    \caption{Training process for (a) the standard BP, (b) the greedy local training method and (c) the proposed local training method with the backward link. $l$ is the propagation length of the errors. $\delta_{l}^{n}$ and $\delta_{g}^{n}$ are the local and global error of the n-th hidden layer, respectively. $W_{c}^{n}$ and $W^{n+1}$ are the weights of the local classifier and the (n+1)-th hidden layer, respectively.}
    \label{fig:1}
\end{figure*}

The remainder of this work consists of the following sections. Section II describes the details of the proposed local training algorithm. Section III discusses the experimental results in various networks. Section IV analyzes the computational costs. Section V discusses performance trade-off and concludes this work.

\section{Method}
The backpropagation algorithm has been the standard training method in deep learning. The training process is depicted in Fig. \ref{fig:1} (a). Assume that a network consists of N layers. In the forward pass, the n-th hidden layer receives an input $x^{n}$ and computes its activations $y^{n}$  by
\begin{equation}
    y^{n}=f(z^{n})=f(W^{n}x^{n}+b^{n})
\end{equation}
where $W^{n}$ is the weight matrix of this layer, $b^{n}$ is the bias vector, $z^{n}$ is the synaptic input, and $f$ is the non-linear activation function, such as Rectified Linear Units (ReLUs). Instead of classifying outputs at the final layer, local training attaches a classifier to the end of each pre-defined local module of the network and predicts the local outputs. The errors generated locally are used to update the module parameters. We refer to the conventional local training method as the greedy local training (GLL) hereafter \cite{RN76}. In the GLL setting, as depicted in Fig. \ref{fig:1} (b), the error flow is prohibited from entering the preceding module, leading to independent training processes. In the backward pass, at the n-th hidden layer, the errors are computed by
\begin{equation}
    \delta^{n} = \frac{\partial L}{\partial z^{n}} = f^{'}(z^{n})\odot (W^{n+1})^{T} \delta^{n+1}
\end{equation}
where $L$ is the loss function, $\odot$ is the element-wise multiplication. The weight gradients are calculated by
\begin{equation}
    \Delta W^{n} = \delta^{n} \otimes x^{n}
\end{equation}
where $\otimes$ is the outer product.

In the proposed local training method, as illustrated in Fig. \ref{fig:1} (c), the errors are propagated backward from one module to its predecessor. Information carried by the error signals can thus be passed between modules. Instead of only focusing on improving the local classification performance, the current module updates its parameters accordingly to improve the performance of subsequent modules. However, to preserve the computational benefits of local training, the errors are restricted from traveling all the way down the module. We define the propagation length of the errors in a module as a hyperparameter, denoted as $l$. For example, $l=2$ is shown in Fig. \ref{fig:1} (c). In this particular case, the greedy local training method corresponds to $l=0$. Any layers located in the propagation range, e.g., the layers in green color in Fig. \ref{fig:1} (c), receive two error contributions we term as local and global contributions, respectively, when updating their errors. These layers can be unfolded into two identical parallel branches that share the same parameters and synchronize parameter updates resulting from different error contributions. If the n-th layer is the last layer of the module, the errors are derived as
\begin{align}
    \delta^{n} &= \delta_{l}^{n} + \delta_{g}^{n} \\
    &= f^{'}(z^{n})\odot [\alpha (W_{c}^{n})^{T} \delta_{c}^{n} + (1-\alpha)(W^{n+1})^{T} \delta^{n+1}]
\end{align}
where $\delta_{l}^{n}$ is the local error vector derived from the errors $\delta_{c}^{n}$ of the local classifier, $\delta_{g}^{n}$ is the global error vector derived from the errors $\delta^{n+1}$ of the next hidden layer, $W_{c}^{n}$ is the weights of the local classifier, and $\alpha$ is the weighting factor. For the layers located before, the errors are derived as
\begin{align}
    \delta^{i} &= \delta_{l}^{i} + \delta_{g}^{i} \\
    &= f^{'}(z^{i})\odot [(W^{i+1})^{T}(\alpha \delta_{l}^{i+1} + (1-\alpha)\delta_{g}^{i+1})]
\end{align}
The unfolding creates two separate network paths and decouples the two adjacent modules so that training can be performed independently.

\section{Classification experiments and results}
\subsection{Experiment setup}
The proposed training method was applied to train three types of widely-used CNNs, namely, AlexNet \cite{RN27}, VGG16 \cite{SimonyanZ14a}, and ResNet \cite{RN42}. The network structures are adapted from the original implementations. In all the networks, convolution operations are performed with a kernel size of 3×3, a stride of 1, and a padding of 1. In ResNets, the residual block consists of a stack of two convolution layers with an identity or down-sampling shortcut. The down-sampling happens at the first residual block where the number of channels doubles and is performed by a 1×1 convolution with a stride of 2. Two types of local classifiers are studied: a simple fully-connected (FC) layer and a convolutional network. The local convolutional network consists of one 3×3 convolutional layer and two FC layers with a hidden feature size of 128.  CIFAR10 and CIFAR100 datasets are used to evaluate the classification performance of the proposed training method \cite{RN84}.
For local training, the networks are split into multiple modules. The number of modules, denoted as K, is chosen as a multiple of 2, varying from 2 to 16 (8 in AlexNet). In ResNets, the residual block is treated as a basic unit (or layer) to preserve the residual structure. We divide the network evenly so that each local module consists of the same number of layers. If the number of basic layers is not divisible by K, we assign one more layer to lower modules. For example, in ResNet110, the number of basic layers is 55, and it can be split into $\{4\}\times7+\{3\}\times9$ for $K=16$ modules. The backward propagation length of the errors, $l$, is adjusted from 0 to 4. The weighting factor $\alpha$ is selected from $\{0,\ 0.25,\ 0.5,\ 0.75,\ 1\}$ and optimized in each network for each classification task.
All the experiments are conducted in the Pytorch framework. A cross-entropy loss function is used at each local classification layer. Stochastic gradient descent (SGD) is used as the optimizer. The momentum is set to 0.9, and the weight decay is set to 0.0001 for VGG16 and 0.0005 for ResNets, respectively. The batch size is set to 512 in all the experiments.  Training is run for 100 epochs in AlexNet, 150 epochs in VGG16, and 200 epochs in ResNets. Learning rates and their schedules are optimized in each network for each classification task. Dropout is only applied in fully-connected layers with the probability of 0.5 \cite{RN34}. ReLU is used as the non-linear function, before which batch normalization was applied.

\subsection{Classification results with local linear classifiers}
CIFAR10 is a collection of RGB frame images from different objects of 10 classes. It is divided into 50,000 training images and 10,000 testing images. We apply the four CNNs with different local training configurations to classify the dataset. Experiments on the impact of different propagation lengths and the number of modules are conducted. Learning rates are optimized in each network, which are 0.01 in AlexNet, 0.01 in VGG16, 0.5 in ResNet32, 0.3 in ResNet110, respectively. 
The average classification errors in 5 trials are presented in Table \ref{tab:1}. The errors resulting from the standard BP are also reported. The number of local modules is changed from 2 to 16. The maximum propagation length is limited by the number of layers inside each local module. It can be observed that the classification performance of the conventional method degrades rapidly with the number of local modules. This is because in the conventional GLL method, without backward connection, local modules only focus on performing their classification task and ignore the demand from the next module, thus causing information loss while updates progress along with the network \cite{wang2021revisiting}. Our method provides a pathway for error signals to travel between modules, therefore retaining useful information to be used by the next modules. It outperforms the GLL method in all cases. The improvement increases with the propagation length, suggesting that more information is passed between modules. The difference becomes more and more prominent as the number of local modules doubles. For example, in ResNet32, with $K=16$ and $l=1$, our method surpasses the conventional method by 4.00\% in accuracy. 

\begin{table}[t]
\centering
\caption{CLASSIFICATION ERRORS ON CIFAR10 AND CIFAR100 DATASETS WHEN LINEAR CLASSIFIERS ARE APPLIED. ERRORS ARE PERCENTAGES. K IS THE NUMBER OF LOCAL MODULES. GLL REFERS TO THE CONVENTIONAL GREEDY LOCAL TRAINING.}
\label{tab:1}
\begin{tabular}{|cccccc|}
\hline
\multicolumn{6}{|l|}{\textbf{CIFAR10}}                                                                                                                                                                                                                                              \\ \hline
\multicolumn{1}{|c|}{Network}                                                                                & \multicolumn{1}{c|}{Method}       & \multicolumn{1}{c|}{K=2}            & \multicolumn{1}{c|}{K=4}            & \multicolumn{1}{c|}{K=8}            & K=16           \\ \hline
\multicolumn{1}{|c|}{\multirow{3}{*}{\begin{tabular}[c]{@{}c@{}}AlexNet\\    \\ (BP: 11.44)\end{tabular}}}   & \multicolumn{1}{c|}{\textit{GLL}} & \multicolumn{1}{c|}{12.43}          & \multicolumn{1}{c|}{12.44}          & \multicolumn{1}{c|}{13.89}          &                \\ \cline{2-6} 
\multicolumn{1}{|c|}{}                                                                                       & \multicolumn{1}{c|}{\textit{l=1}} & \multicolumn{1}{c|}{12.24}          & \multicolumn{1}{c|}{12.43}          & \multicolumn{1}{c|}{\textbf{12.45}} &                \\ \cline{2-6} 
\multicolumn{1}{|c|}{}                                                                                       & \multicolumn{1}{c|}{\textit{l=2}} & \multicolumn{1}{c|}{\textbf{11.93}} & \multicolumn{1}{c|}{\textbf{11.96}} & \multicolumn{1}{c|}{}               &                \\ \hline
\multicolumn{1}{|c|}{\multirow{3}{*}{\begin{tabular}[c]{@{}c@{}}VGG16\\    \\ (BP: 7.41)\end{tabular}}}      & \multicolumn{1}{c|}{\textit{GLL}} & \multicolumn{1}{c|}{10.93}          & \multicolumn{1}{c|}{11.34}          & \multicolumn{1}{c|}{13.65}          & 18.02          \\ \cline{2-6} 
\multicolumn{1}{|c|}{}                                                                                       & \multicolumn{1}{c|}{\textit{l=1}} & \multicolumn{1}{c|}{10.68}          & \multicolumn{1}{c|}{11.18}          & \multicolumn{1}{c|}{11.74}          & \textbf{12.72} \\ \cline{2-6} 
\multicolumn{1}{|c|}{}                                                                                       & \multicolumn{1}{c|}{\textit{l=2}} & \multicolumn{1}{c|}{\textbf{10.54}} & \multicolumn{1}{c|}{\textbf{10.58}} & \multicolumn{1}{c|}{\textbf{10.82}} &                \\ \hline
\multicolumn{1}{|c|}{\multirow{5}{*}{\begin{tabular}[c]{@{}c@{}}ResNet32\\    \\ (BP: 7.51)\end{tabular}}}   & \multicolumn{1}{c|}{\textit{GLL}} & \multicolumn{1}{c|}{10.35}          & \multicolumn{1}{c|}{15.02}          & \multicolumn{1}{c|}{20.59}          & 24.21          \\ \cline{2-6} 
\multicolumn{1}{|c|}{}                                                                                       & \multicolumn{1}{c|}{\textit{l=1}} & \multicolumn{1}{c|}{10.02}          & \multicolumn{1}{c|}{14.27}          & \multicolumn{1}{c|}{17.53}          & \textbf{20.21} \\ \cline{2-6} 
\multicolumn{1}{|c|}{}                                                                                       & \multicolumn{1}{c|}{\textit{l=2}} & \multicolumn{1}{c|}{9.03}           & \multicolumn{1}{c|}{12.92}          & \multicolumn{1}{c|}{\textbf{16.23}} &                \\ \cline{2-6} 
\multicolumn{1}{|c|}{}                                                                                       & \multicolumn{1}{c|}{\textit{l=3}} & \multicolumn{1}{c|}{9.22}           & \multicolumn{1}{c|}{10.65}          & \multicolumn{1}{c|}{}               &                \\ \cline{2-6} 
\multicolumn{1}{|c|}{}                                                                                       & \multicolumn{1}{c|}{\textit{l=4}} & \multicolumn{1}{c|}{\textbf{9.18}}  & \multicolumn{1}{c|}{\textbf{10.63}} & \multicolumn{1}{c|}{}               &                \\ \hline
\multicolumn{1}{|c|}{\multirow{5}{*}{\begin{tabular}[c]{@{}c@{}}ResNet110\\    \\ (BP: 6.63)\end{tabular}}}  & \multicolumn{1}{c|}{\textit{GLL}} & \multicolumn{1}{c|}{8.55}           & \multicolumn{1}{c|}{13.19}          & \multicolumn{1}{c|}{14.97}          & 17.9           \\ \cline{2-6} 
\multicolumn{1}{|c|}{}                                                                                       & \multicolumn{1}{c|}{\textit{l=1}} & \multicolumn{1}{c|}{8.34}           & \multicolumn{1}{c|}{12.6}           & \multicolumn{1}{c|}{14.92}          & 17.51          \\ \cline{2-6} 
\multicolumn{1}{|c|}{}                                                                                       & \multicolumn{1}{c|}{\textit{l=2}} & \multicolumn{1}{c|}{7.99}           & \multicolumn{1}{c|}{12.56}          & \multicolumn{1}{c|}{14.64}          & 16.18          \\ \cline{2-6} 
\multicolumn{1}{|c|}{}                                                                                       & \multicolumn{1}{c|}{\textit{l=3}} & \multicolumn{1}{c|}{\textbf{7.96}}  & \multicolumn{1}{c|}{\textbf{12.38}} & \multicolumn{1}{c|}{14.38}          & \textbf{15.62} \\ \cline{2-6} 
\multicolumn{1}{|c|}{}                                                                                       & \multicolumn{1}{c|}{\textit{l=4}} & \multicolumn{1}{c|}{8.25}           & \multicolumn{1}{c|}{12.54}          & \multicolumn{1}{c|}{\textbf{13.62}} &                \\ \hline
\multicolumn{6}{|l|}{\textbf{CIFAR100}}                                                                                                                                                                                                                                             \\ \hline
\multicolumn{1}{|c|}{Network}                                                                                & \multicolumn{1}{c|}{Method}       & \multicolumn{1}{c|}{K=2}            & \multicolumn{1}{c|}{K=4}            & \multicolumn{1}{c|}{K=8}            & K=16           \\ \hline
\multicolumn{1}{|c|}{\multirow{5}{*}{\begin{tabular}[c]{@{}c@{}}ResNet32\\    \\ (BP: 29.68)\end{tabular}}}  & \multicolumn{1}{c|}{\textit{GLL}} & \multicolumn{1}{c|}{34.02}          & \multicolumn{1}{c|}{42.44}          & \multicolumn{1}{c|}{48.32}          & 51.75          \\ \cline{2-6} 
\multicolumn{1}{|c|}{}                                                                                       & \multicolumn{1}{c|}{\textit{l=1}} & \multicolumn{1}{c|}{33.60}          & \multicolumn{1}{c|}{41.92}          & \multicolumn{1}{c|}{45.12}          & \textbf{46.76} \\ \cline{2-6} 
\multicolumn{1}{|c|}{}                                                                                       & \multicolumn{1}{c|}{\textit{l=2}} & \multicolumn{1}{c|}{32.56}          & \multicolumn{1}{c|}{38.95}          & \multicolumn{1}{c|}{\textbf{42.59}} &                \\ \cline{2-6} 
\multicolumn{1}{|c|}{}                                                                                       & \multicolumn{1}{c|}{\textit{l=3}} & \multicolumn{1}{c|}{32.64}          & \multicolumn{1}{c|}{36.61}          & \multicolumn{1}{c|}{}               &                \\ \cline{2-6} 
\multicolumn{1}{|c|}{}                                                                                       & \multicolumn{1}{c|}{\textit{l=4}} & \multicolumn{1}{c|}{\textbf{32.47}} & \multicolumn{1}{c|}{\textbf{35.99}} & \multicolumn{1}{c|}{}               &                \\ \hline
\multicolumn{1}{|c|}{\multirow{5}{*}{\begin{tabular}[c]{@{}c@{}}ResNet110\\    \\ (BP: 28.33)\end{tabular}}} & \multicolumn{1}{c|}{\textit{GLL}} & \multicolumn{1}{c|}{31.37}          & \multicolumn{1}{c|}{37.22}          & \multicolumn{1}{c|}{41.73}          & 45.23          \\ \cline{2-6} 
\multicolumn{1}{|c|}{}                                                                                       & \multicolumn{1}{c|}{\textit{l=1}} & \multicolumn{1}{c|}{30.85}          & \multicolumn{1}{c|}{37.5}           & \multicolumn{1}{c|}{41.49}          & 44.82          \\ \cline{2-6} 
\multicolumn{1}{|c|}{}                                                                                       & \multicolumn{1}{c|}{\textit{l=2}} & \multicolumn{1}{c|}{30.95}          & \multicolumn{1}{c|}{36.94}          & \multicolumn{1}{c|}{41.34}          & 43.85          \\ \cline{2-6} 
\multicolumn{1}{|c|}{}                                                                                       & \multicolumn{1}{c|}{\textit{l=3}} & \multicolumn{1}{c|}{30.54}          & \multicolumn{1}{c|}{36.37}          & \multicolumn{1}{c|}{41.05}          & \textbf{42.09} \\ \cline{2-6} 
\multicolumn{1}{|c|}{}                                                                                       & \multicolumn{1}{c|}{\textit{l=4}} & \multicolumn{1}{c|}{\textbf{30.28}} & \multicolumn{1}{c|}{\textbf{35.63}} & \multicolumn{1}{c|}{\textbf{38.88}} &                \\ \hline
\end{tabular}
\end{table}

CIFAR100 has the same dataset size as CIFAR10, but it has 100 classes. It is a more challenging classification task, a good benchmark for evaluating the proposed method. We apply ResNet32 and ResNet110 with different local training configurations to classify the dataset. The same experiments are performed to analyze the impact of our method. The average classification errors in 5 trials are presented in Table \ref{tab:1}. The same observations can be made as follows. Increasing the number of local modules degrades the classification performance. The degradation becomes more severe than in the case of CIFAR10. Our method reduces classification errors in all cases. Especially, the reduction becomes significant when the number of local modules is large. For example, in ResNet32, with $K=16$ and $l=1$, our method surpasses the GLL method by 4.99\% in accuracy. Therefore, our method is demonstrated to be consistent in improving classification performance of local training algorithms. Additionally, ResNet32 benefits more from the backward dependency than ResNet110, as the improvement is more evident. Particularly, when $K=4$, ResNet32 achieves better accuracy on CIFAR10 and similar accuracy on CIFAR100.

\subsection{Classification results with local convolutional classifiers}
Increasing the complexity of the local classifiers helps train the local modules to produce features containing useful information for the subsequent modules \cite{pmlr-v119-belilovsky20a, RN76}. We perform experiments in ResNet32 and ResNet110 with local convolutional classifiers consisting of one convolutional layer and two FC layers. The classification results on CIFAR10 are obtained in Table \ref{tab:2}. We include two state-of-the-art local training methods for comparisons. The decoupled greedy learning (DGL) was proposed to parallelize module computations in both forward and backward pass by means of local replay buffers and classifiers \cite{pmlr-v119-belilovsky20a}. Wang et al. analyzed the drawbacks of local training methods from information perspectives and proposed an information propagation (InfoPro) loss in the local optimization objective to preserve information in local modules \cite{wang2021revisiting}. The same local classifiers are adopted for fair comparisons. It is clear that in all the methods, the accuracy drops as the number of local modules increases. Overall, our method achieves the best classification performance. Compared with the InfoPro method, our method shows notably better accuracy in ResNet32 and slight increase in ResNet110. The InfoPro method relies on a complex local network architecture composed of a convolutional classifier and a convolutional decoder and introduces four additional hyperparameters for tuning. The decoder is built on a bilinear interpolation layer and two convolutional layers to reconstruct the input image locally. In contrast, our method has much lower complexity. 
\begin{table}[t]
\centering
\caption{CLASSIFICATION ERRORS ON CIFAR10 DATASET WHEN CONVOLUTIONAL CLASSIFIERS ARE APPLIED. ERRORS ARE PERCENTAGES. K IS THE NUMBER OF LOCAL MODULES. GLL REFERS TO THE CONVENTIONAL GREEDY LOCAL TRAINING.}
\label{tab:2}
\begin{tabular}{|c|c|c|c|c|c|}
\hline
Network                                                                               & Method                   & K=2           & K=4           & K=8            & K=16           \\ \hline
\multirow{7}{*}{\begin{tabular}[c]{@{}c@{}}ResNet32\\    \\ (BP: 6.94)\end{tabular}}  & \textit{GLL}             & 7.96          & 10.61         & 13.61          & 14.97          \\ \cline{2-6} 
                                                                                      & \textit{DGL{[}14{]}}     & 8.69          & 11.48         & 14.17          & 16.22          \\ \cline{2-6} 
                                                                                      & \textit{InfoPro{[}15{]}} & 8.13          & 8.64          & 11.4           & 14.23          \\ \cline{2-6} 
                                                                                      & \textit{l=1}             & 7.44          & 9.65          & 11.7           & \textbf{12.40} \\ \cline{2-6} 
                                                                                      & \textit{l=2}             & 7.17          & 8.66          & \textbf{10.60} &                \\ \cline{2-6} 
                                                                                      & \textit{l=3}             & 7.13          & 7.95          &                &                \\ \cline{2-6} 
                                                                                      & \textit{l=4}             & \textbf{6.97} & \textbf{7.55} &                &                \\ \hline
\multirow{7}{*}{\begin{tabular}[c]{@{}c@{}}ResNet110\\    \\ (BP: 6.27)\end{tabular}} & \textit{GLL}             & 7.45          & 10.07         & 11.87          & 12.99          \\ \cline{2-6} 
                                                                                      & \textit{DGL{[}14{]}}     & 7.70          & 10.50         & 12.46          & 13.80          \\ \cline{2-6} 
                                                                                      & \textit{InfoPro{[}15{]}} & 7.01          & 7.96          & 9.40           & 10.78          \\ \cline{2-6} 
                                                                                      & \textit{l=1}             & 6.47          & 8.65          & 9.95           & 10.77          \\ \cline{2-6} 
                                                                                      & \textit{l=2}             & 6.39          & 7.96          & 9.56           & 10.61          \\ \cline{2-6} 
                                                                                      & \textit{l=3}             & \textbf{6.36} & 7.92          & 9.29           & \textbf{9.75}  \\ \cline{2-6} 
                                                                                      & \textit{l=4}             & 6.38          & \textbf{7.79} & \textbf{9.25}  &                \\ \hline
\end{tabular}
\end{table}

\section{Computational costs}
Local training algorithms are the favored alternative approach to the standard BP because of their low memory cost and excellent model parallelization. In this section, we will analyze the computational cost of the proposed method and compare it with other methods.
\subsection{GPU memory cost}
The greedy local training splits networks into gradient-isolated modules, eliminating the necessity to save the whole computational graph and intermediate tensors and leading to reduced memory cost. While our method allows errors to flow between modules and hence imposes the backward dependency, we limit the propagation length within the local module, reduce the extent of dependency, and preserve the advantage of low memory usage.

\begin{figure}[b]
    \centering
    \includegraphics[width=\linewidth]{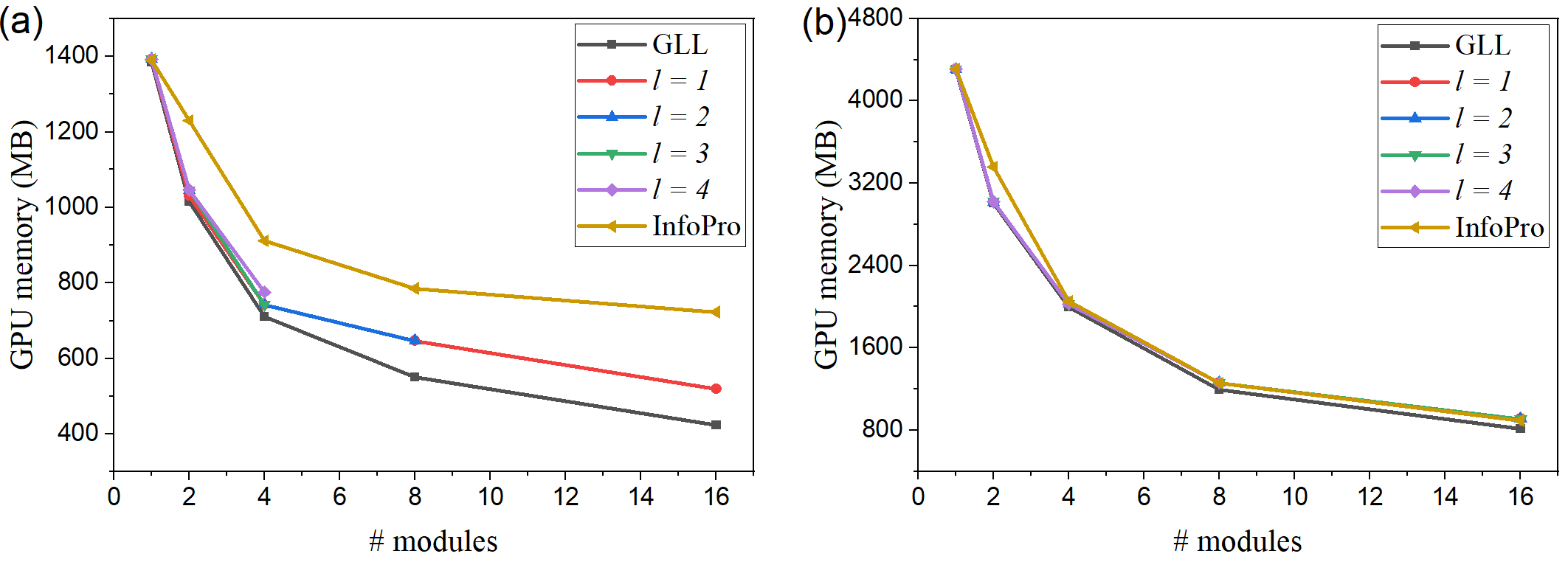}
    \caption{GPU memory cost of training (a) ResNet32 and (b) ResNet110 with local convolutional classifiers on a single GPU. The standard BP corresponds to 1 local module.}
    \label{fig:2}
\end{figure}

We measure the maximum GPU memory occupied by ResNets with local convolutional classifiers under different settings on a single Nvidia Titan RTX GPU. The measurement is done with the commonly-used command \textit{max\_memory\_allocated} in Pytorch \cite{wang2021revisiting, RN57}. The results of training ResNet32 and ResNet110 are shown in Fig.\ref{fig:2} (a) and (b), respectively. As expected, the memory cost decreases significantly with the number of local modules. Compared with the standard BP, local training can lead up to 69\% in ResNet32 and 81\% in ResNet110. Applying our method introduces a small overhead, which is 23\% at maximum in ResNet32 and 12\% in ResNet110. In most cases, the overhead is smaller than 5\%. Moreover, due to the complexity of the local network, the InfoPro method consumes much more GPU memory in ResNet32 with an overhead of 71\% when the number of modules is 16. However, the overhead becomes less in ResNet110 because the size of the local network is much smaller than the local module. Therefore, compared with the GLL method, our method causes a smaller overhead in GPU memory cost than the InfoPro method.
\subsection{Runtime cost on multiple GPUs}
\begin{figure}
    \centering
    \includegraphics[width=0.6\linewidth]{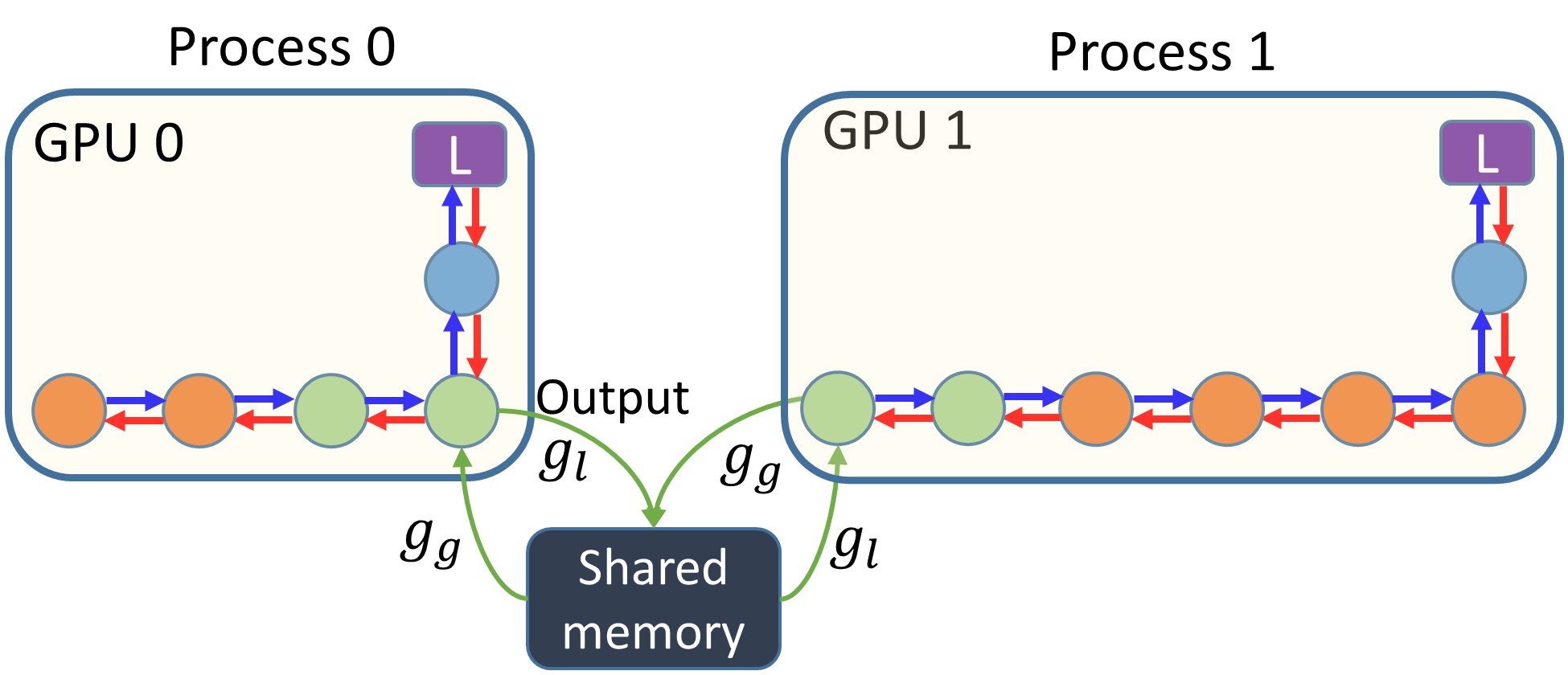}
    \caption{Multi-GPU implementation of the proposed local training method. $g_{l}$ and $g_{g}$ are the local gradients and global gradients, respectively. The green circles indicate the hidden layers with the backward link.}
    \label{fig:3}
\end{figure}

\begin{figure}
    \centering
    \includegraphics[width=\linewidth]{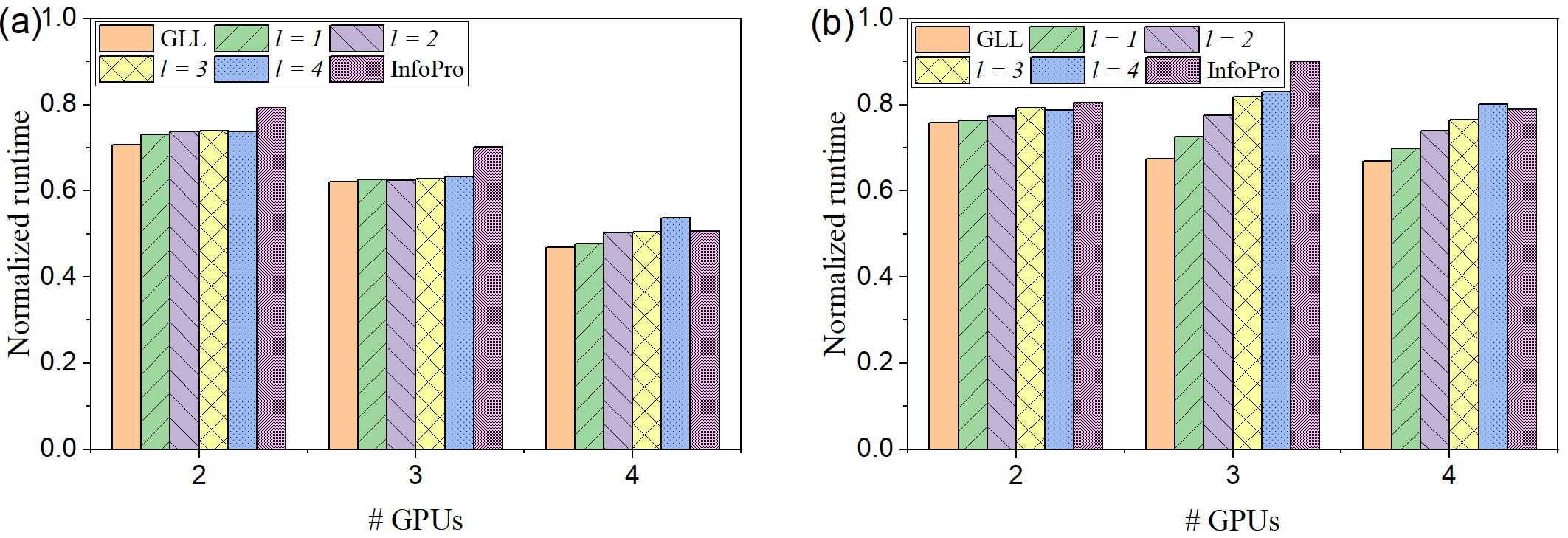}
    \caption{Runtime comparison among different local training methods in ResNet110 on multiple GPUs for two different GPU models: (a) Nvidia Tesla P100 and (b) Nvidia Tesla V100. The runtime is normalized over the standard BP.}
    \label{fig:4}
\end{figure}

Local training algorithms have inherent model parallelization that enables the parallel execution of the forward pass of a module and the backward pass of its previous module. To leverage the parallelization, we can implement networks on multiple GPUs, each running a local module as a single process. Each GPU receives the inputs from the previous one except the very first one that gets the inputs from the dataset. The inputs are saved in shared memory. In this way, each GPU performs training in the local module independently, leading to reduced training runtime.
Similarly, our training method can be implemented on multiple GPUs to achieve training acceleration. The implementation is illustrated in Fig. \ref{fig:3}. The example shows an 8-layer network split into two local modules with an error propagation length of 2. Each GPU trains one module. We duplicate the last two layers (in green) of the first module into the second module to compute their gradients resulting from the global errors on the same GPU. The local gradients are computed on GPU0. Then the final gradients of these two layers are synchronized across GPUs by exchanging the gradients through shared memory. The gradients from the two layers on the same GPU can be concatenated and sent out to the shared memory at once to reduce the communication time. In this way, training on different GPUs is performed independently with an overhead of gradient communication time.
We utilize the multiprocessing package in Pytorch to create multiple subprocesses and necessary shared memories \cite{pmlr-v119-belilovsky20a, osti_10074580}. We split ResNet110 into K modules and measure the simulation runtime on K GPUs. The convolutional classifier is used in local modules. Runtime is measured on two different types of GPU models, namely Nvidia Tesla P100 and V100, as presented in Fig. \ref{fig:4} (a) and (b), respectively. The runtime is normalized over the standard BP. From Fig. \ref{fig:4}, we can tell that GPU models affect the local training speed. The GLL method significantly reduce the runtime by up to 53\% on P100 GPUs and 23\% on V100 GPUs compared with the standard BP, respectively. For the P100 model, the runtime is improved as the number of GPUs increases, while little or even no runtime improvement can be observed with more than two GPUs of the V100 models. The difference could be attributed to the GPU processing speed and the communication time between GPUs and the shared memory. More importantly, our method runs faster than the InfoPro method and causes a small runtime overhead, which are less than 5\% for the P100 model and 10\% for the V100 model in most of the cases, respectively.

\section{Discussion and conclusion}
\begin{figure}[b]
    \centering
    \includegraphics[width=0.6\linewidth]{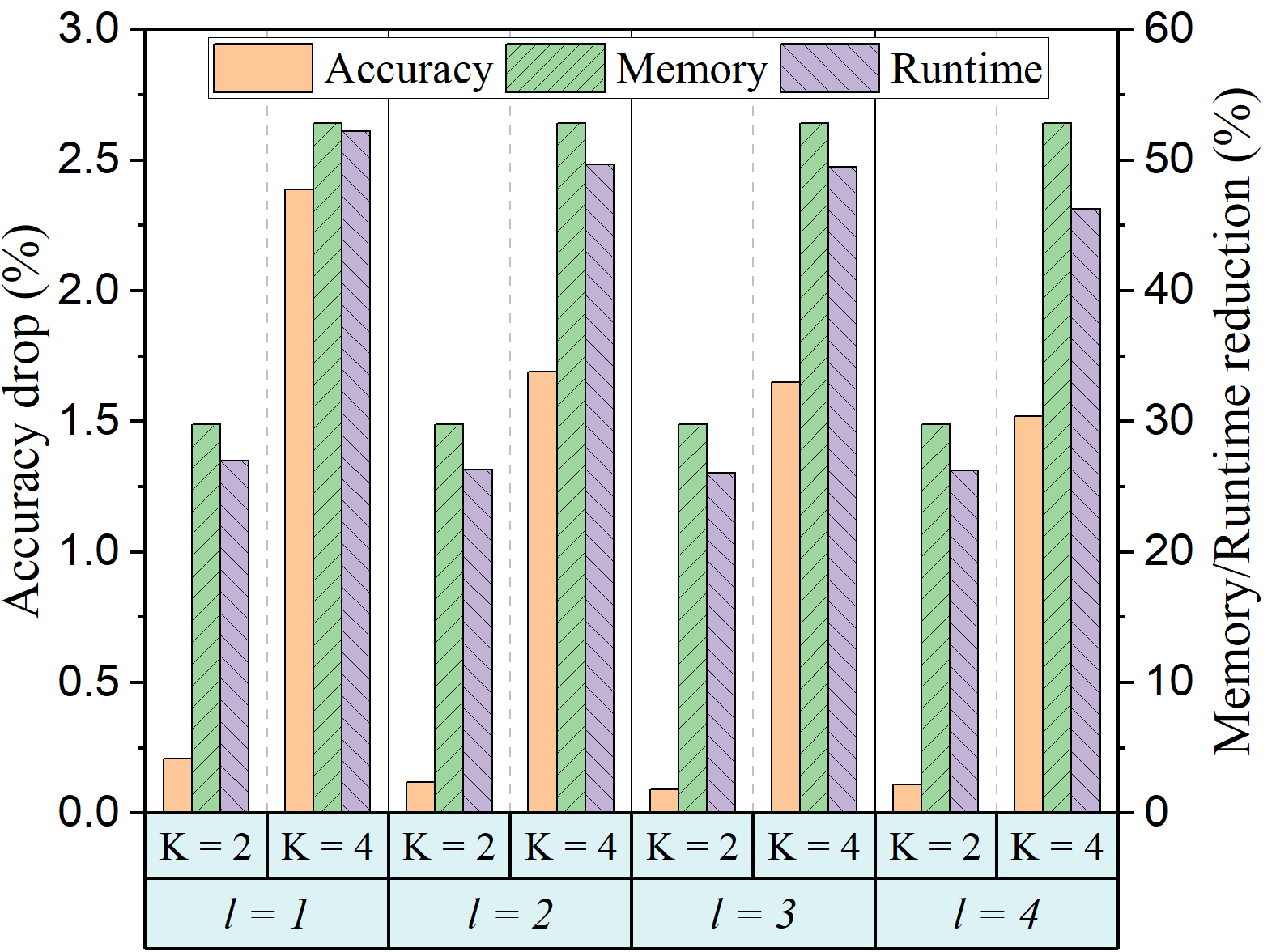}
    \caption{Performance trade-off resulting from different number of local modules for the error propagation length changing from 1, 2, 3, to 4. Accuracy drop, GPU memory and runtime reduction are computed with the standard BP as the baseline. Results are obtained from ResNet110 on CIFAR10.}
    \label{fig:5}
\end{figure}

The proposed BackLink local training algorithm is demonstrated to produce accuracy close to the standard BP, outperforming the state-of-the-art local training methods. It has shown a great advantage in lowering GPU memory cost and runtime by increasing the number of local modules. However, classification performance is also impacted. Thus, we present the performance trade-off among the accuracy drop, GPU memory and runtime, as shown in Fig. \ref{fig:5}. For the same propagation length $l$, increasing the number of local modules incurs more accuracy drop, but leads to higher memory and runtime reduction. Specifically, in the case of $l=2$, when the number of local modules doubles, 1.57\% drop in accuracy is caused, while around 23\% more reduction in memory and runtime are achieved. On the other hand, increasing the propagation length from 1 to 4 leads to improvement in accuracy, i.e. 0.87\%, with negligible overhead, since the memory reduction remains unchanged and the runtime is slightly affected (< 6\%). Accuracy improevement is more significant with a larger K, as shown in Table \ref{tab:1} and Table \ref{tab:1}.
In this work, we have demonstrated a new local training algorithm that introduces a backward dependency between local modules by allowing errors to flow between them and guides information to travel along with the network from top to bottom. The proposed algorithm tackles the shortcoming of the conventional local training algorithm that loses information while features are propagated forward along with the network. The backward dependency ensures that the preceding modules preserve enough information in the output features to be utilized by the subsequent modules. The network can achieve better global classification performance instead of only focusing on improving individual local performance. Extensive experiments on different types of CNNs and benchmark datasets have demonstrated the effectiveness of the proposed algorithm and considerable improvement in classification performance. For example, in ResNet32 with 16 local modules, our method surpasses the conventional greedy local training method by 4.00\% and the recent InfoPro method by 1.83\% in accuracy on CIFAR10, respectively. In order to preserve the computational advantages of local training algorithms, we lighten the backward dependency between modules by restricting the propagation length of the errors within a module. As a result, computational advantages of local training in GPU memory and multi-GPU simulation runtime, are marginally impacted, while significant reductions are still achieved in comparison to the standard BP. Our method can lead up to a 79\% reduction in memory cost and 53\% in simulation runtime in ResNet110 compared to the standard BP. Therefore, we believe that our algorithm brings new possibilities for developing more efficient and biologically plausible deep learning algorithms.


\end{document}